# INDOOR 3D VIDEO MONITORING USING MULTIPLE KINECT DEPTH-CAMERAS

M. Martínez-Zarzuela[1], M. Pedraza-Hueso, F.J. Díaz-Pernas,

D. González-Ortega, M. Antón-Rodríguez

[1]Departmento de Teoría de la Señal y Comunicaciones e Ingeniería Telemática
University of Valladolid
`marmar@tel.uva.es`

## ABSTRACT

*This article describes the design and development of a system for remote indoor 3D monitoring using an undetermined number of Microsoft® Kinect sensors. In the proposed client-server system, the Kinect cameras can be connected to different computers, addressing this way the hardware limitation of one sensor per USB controller. The reason behind this limitation is the high bandwidth needed by the sensor, which becomes also an issue for the distributed system TCP/IP communications. Since traffic volume is too high, 3D data has to be compressed before it can be sent over the network. The solution consists in self-coding the Kinect data into RGB images and then using a standard multimedia codec to compress color maps. Information from different sources is collected into a central client computer, where point clouds are transformed to reconstruct the scene in 3D. An algorithm is proposed to merge the skeletons detected locally by each Kinect conveniently, so that monitoring of people is robust to self and inter-user occlusions. Final skeletons are labeled and trajectories of every joint can be saved for event reconstruction or further analysis.*

## 1. INTRODUCTION

A system for remote people monitoring can be employed in a large amount of useful applications, such as those related to security and surveillance [1], human behavior analysis [2] and elderly people or patient health care [3] [4]. Due to their significance, human body tracking and monitoring are study fields in computer vision that have always attracted the interest of researchers [5][6]. As a result, many technologies and methods have been proposed. Computer vision techniques are becoming increasingly sophisticated, aided by new acquisition devices and low-cost hardware data processing capabilities.

The complexity of the proposed methods can significantly depend on the way the scene is acquired. An important requirement is to achieve a fine human silhouette segmentation. State-of-the-art technologies are really good at this task. Apart from the techniques that use markers attached to the human body, tracking operations are carried out mainly in two ways, from 2D information or 3D information [7] [8]. On the one hand, 2D body tracking is presented as the classic solution; a region of interest is detected within a 2D image and processed. Because of the use of silhouettes, this method suffers occlusions. On the other hand, advanced body tracking and pose estimation is currently being carried out by means of 3D cameras, such as binocular, Time-of-Flight (ToF) or consumer depth-cameras like Microsoft(R) Kinect [9]. The introduction of low-cost depth sensors has pushed up the development of new systems based on robust segmentation and tracking of human skeletons. The number of applications built on top of depth-sensor devices is rapidly increasing. However, most of these new systems are aimed to track only one or two people thus have only direct application on videogames or human-computer interfaces.

There are some limitations to address in order to build a remote space monitoring system using consumer depth-cameras, and only a few separate efforts have been done to address these limitations. Even so, those developments do not pursue building a remote monitoring system, but covering part of the limitations in which we are also interested for our system. On the one hand, Kinect devices can capture only a quite small area, covering accurately distances only up to 3.5 meters [9]. There are proposals which allow to make a 3D reconstruction of spaces and objects using Kinect [10], but in them every capturing device has to be connected to the same computer. Apart from that, these solutions cannot merge skeletons information from different Kinects. The first limitation is significant, since only two or three devices can be connected to a single computer, due to the high USB bandwidth consumption of these cameras. There is another proposal that allows to send data over a network [11]. However, this application uses Microsoft SDK [9], so it only works under Windows operating system.

The 3D monitoring system presented in this paper addresses these limitations and allows using an undetermined number of Microsoft® Kinect cameras, connected to an undetermined number of computers running any operating system (Windows, Linux, Mac), to monitor people in a large space remotely. The system codes the 3D information (point clouds representing the scene, human skeletons and silhouettes) acquired by each camera, so that bandwidth requirements for real-time monitoring are met. The information coming from different devices is synchronized. Point clouds are combined to reconstruct the scene in 3D and human skeletons and silhouettes information coming from different cameras are merged conveniently to build a system robust to self-user or inter-user occlusions. The proposed system uses low-cost hardware and open source software libraries, which makes its deployment affordable for many applications under different circumstances.

## 2. TOOLS AND METHODS

### 2.1 Consumer depth-cameras

For 3D scene acquisition, a number of devices can be used. For computer vision techniques, we can distinguish between passive and active cameras. The first include stereo devices, simulating the left and right eye in human vision: the images coming from each camera in the device are combined to generate a disparity map and reconstruct depth information [7]. In this category, some other proposals in which several passive cameras are disposed around the person or object to be reconstructed can be included. The second option consists in using an active device such as a ToF (Time of Flight) camera or newer consumer depth-cameras like Microsoft® Kinect or ASUS® Xtion. Despite their high depth precision, ToF cameras are expensive and provide very low resolutions. On the other side, consumer depth-cameras provide resolutions starting at 640x480 px and 30 fps at very affordable prices.

We will pay special attention to Kinect cameras, since they are the chosen devices for the proposed system. Microsoft® Kinect emits a structured infrared pattern of points over its field of view, which is then captured by a sensor and employed to estimate the depth of every projected point of the scene. Although Kinect was initially devised only for computer games, the interest of the computer vision community rapidly made it possible to use the device for general purpose from a computer, even before the Microsoft® official Kinect SDK was available[9]. There is a wide variety of tools to work with Kinect. A commonly used framework for creating applications is OpenNI [12], which has been developed to be compatible with any commodity depth-camera and, in combination with NiTE middleware, is able to automate tasks for user identifying, feature detection, and basic gesture recognition [13].

## 2.2 Data compression

Consumer depth-cameras generate a large volume of data. This is an important issue, since one of the objectives of the system is the transmission of this information over a network. Therefore, data compression is necessary before sending data to a central computer. There are different ways to compress data. If the data to compress is not multimedia, we can use a zip encoder, which provides lossless compression, but generates large output data and is computationally expensive. For multimedia compression, there are picture encoders such as jpeg, which do not use temporal redundancy. To compress video, there are many encoders like H.264 or VP8. These encoders are able to compress data taking advantage of the temporal redundancy, thus compressed information is suitable to be sent over the network. However, there are not extended codecs to compress depth maps yet. One type of compression codecs used for 3D images, are those used to transmit the 3D television signal, but they are based on the compression of two images (right and left) [14], thus they are not useful for our system, in which 3D information is directly acquired using an active infrared device.

## 2.3 CORBA

A distributed application based on the client-server paradigm does not need to be developed using low level sockets. For the proposed system, a much more convenient approach is to use a middleware such as TAO CORBA, a standard defined by OMG (Object Management Group). This middleware allows using a naming service [15], that avoids the central client to know about the addresses of each one of the servers. The aim of CORBA is hiding the low level complexity algorithms for data transmission over the network from the programmer. It is object-oriented and supports C++, Python, Java, XML, Visual Basic, Ada, C, COBOL, CORBA-Scripting-Language, Lisp, PL/1, Smalltalk and C#. Besides, this middleware is chosen because it is independent of the programming language, so servers could be programmed in Java and a client in C++, for example. It represents a clear advantage over RMI, which can only be programmed in Java. CORBA is also cross platform, so clients and servers can be running on different operating systems. In the proposed system, the servers may be running on Windows computers and the client in a Linux computer or in the opposite way.

## 2.4 PCL: Point Cloud Library

PCL 'Point Cloud Library' [16], is a C++ free open source computer vision library to work with 3D information that can be used in many areas such as robotics. PCL is being developed by a group of researchers and engineers from around the world. There are also many companies such as Toyota or Nvidia working to develop this powerful library [17]. The library contains algorithms for filtering, feature estimating, point cloud registration, and segmentation.

Point clouds can be obtained and stored in 3D raw data files, read from 3D models or 3D cameras such as Kinect. The combination of both technologies, PCL and Kinect, is very convenient for our purpose of monitoring a space with 3D information. The library is comprised of the following modules: filters, features, keypoints, registration, kdtree, octree, segmentation, simple consensus, surface, range image, IO, visualization, common, and search.

## 3. PROPOSED SYSTEM

The main feature of the proposed system is its capability of merging 3D information coming from multiple Kinect devices, including depth information and detected skeletons. This takes place under the client-server model, in which servers are computers with attached devices and the client is one or more central computers responsible for information fusion, tracking, and visualization.

### 3.1 General description

Figure 1 depicts the scheme of the proposed system. A server is a computer where one or more Kinect cameras are connected. The different servers, deployed in a remote space are responsible for capturing the information coming from different regions of the scene. This information is conveniently processed and then sent to the clients. The large amount of information acquired by Kinect devices has to be compressed using different strategies before it can be sent over the network. The central client is in charge of reconstructing the remote space in 3D using PCL library and includes a robust algorithm for merging of multiple detected skeletons. The computer interface can be used to monitor the scene in 3D in real time, label people within it and record specific users movements for further analysis [18] The system is fully scalable to any number of servers and clients, thus to any number of acquiring devices and locations.

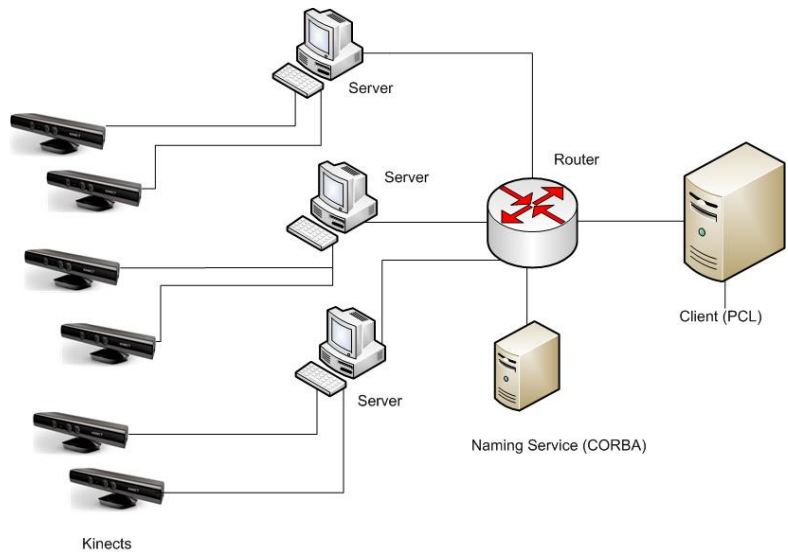

Figure 1. General scheme of the proposed system.

## 3.2 Environment configuration

As explained before, Kinect cameras are devices that emit infrared patterns of light in order to acquire depth information within a limited range. Kinect devices have a view range that covers depth precisely only between 0.5 m and 3.5 m [9]. For this reason, one of the motivations of our system is to expand the covered view by adding several Kinects to the scenario. However, an important issue arises when placing various Kinect cameras in the same place: the infrared pattern emitted by different Kinects can interfere with each other, causing 'holes' in the acquired point clouds. Therefore, we must be careful in the placement of the devices and avoid placing a camera right in front of another one. Figure 2 shows the lack of 3D information around the bookshelves when a second camera is positioned to capture the same region of the environment. There is some interference among the infrared emitted patterns that prevents scene reconstruction.

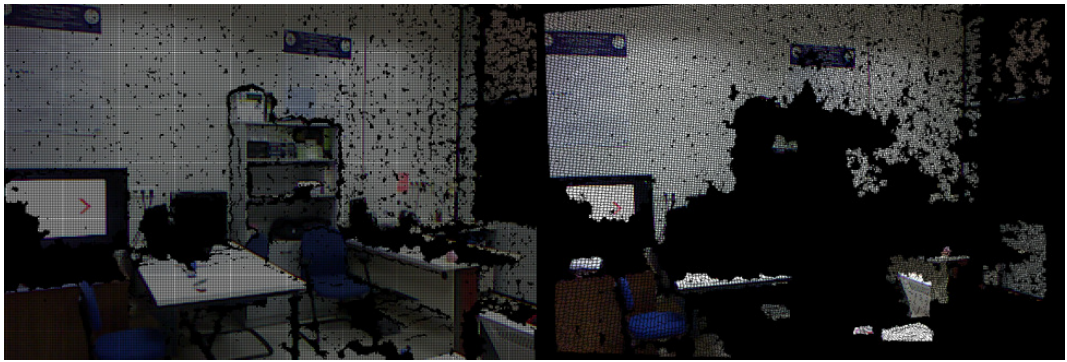

Figure 2. Scene without and with infrared interference.

With the aim of minimizing this interference, we studied which is the best and more flexible camera layout, flexible enough to cover most of the environments. A correct placement of cameras has to take into account not only how to reduce camera interferences, but also favor robust skeleton tracking when a person in the scene walks from the area covered by one camera to the area covered by another one.

A naïve solution consists in placing the cameras in a layout that avoids any possible collision of the cameras' infrared patterns, in a scheme depicted in Figure 3(a). The Kinect camera provides an angle of view of 57.5 degrees, thus we can calculate the distance between two cameras. Following the trigonometric relation in equation (1), for a given depth distance $d$ covered by one Kinect, the next Kinect camera should be at a distance $2h$, being $h \cong 0.55d$

$$\tan(57.5º) = h/d \qquad (1)$$

Under these circumstances, we can merge the point clouds without any information holes. However, the problem behind placing the cameras just like that is that the system loses some scene information, since the space between cameras is not covered by any infrared pattern. The next logical improvement to this layout is placing complementary Kinects in the opposite side, as shown in Figure 3(b). It is important that those Kinects placed across from those placed in first place are laterally shifted, so that we can capture all the regions and users in the scene. There will always be one camera in front of a person and we cover all the possible dark points among the cameras.

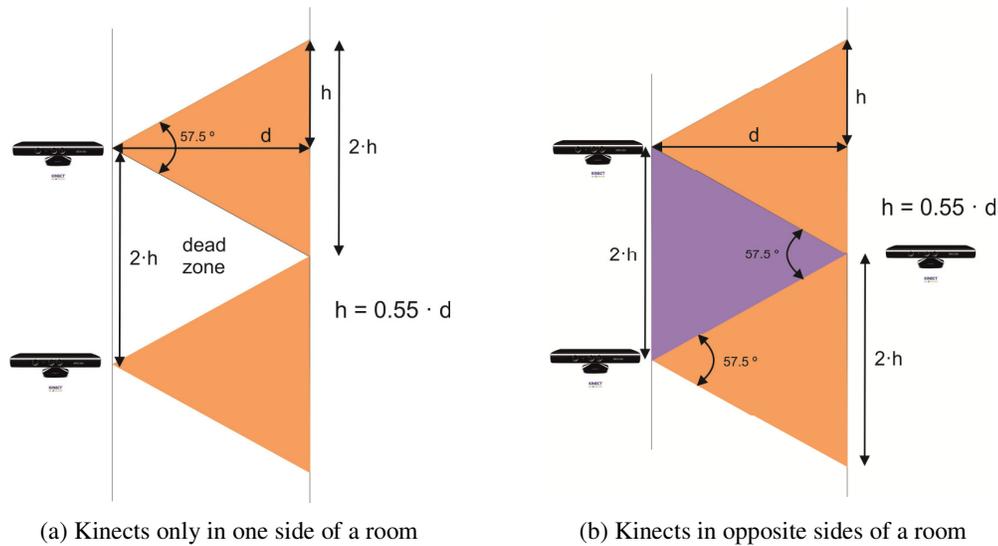

(a) Kinects only in one side of a room     (b) Kinects in opposite sides of a room

Figure 3. Naïve Kinect cameras layout

Although this layout is free from interference, it is not the most appropriate for skeleton tracking. The problem arises when a person passes through the areas covered by two cameras, since there is a non-negligible delay when new skeletons have to be detected and processed. The first camera will lose the user information and the second camera needs a certain time to find the limbs and joints of the new user that has appeared.

For this reason, the camera layout for the final proposed system does not try to remove camera interference completely, but minimize it. Despite the fact that we want cameras do not interfere, we want the areas covered by different cameras to be overlapped enough so that a new skeleton can be detected fast enough. A good trade-off for region overlapping is the one that covers a distance of 0.5 m among the cameras, as in the scheme shown in Figure 4. With regard to the distance from the cameras to the floor, we recommend holding them at 1.85 m and slightly tilted down for faster skeleton detection. In general, Kinect cameras are recommended to be placed at the height of the hips. However, by placing them at the proposed height, increases the number of items that can be covered in the scene.

It is worth mentioning that this is not the only possible cameras layout for the system to work, but only the most recommended one. Actually, the system is flexible enough to work with any cameras layout configuration. The system can be adapted to any room configuration, no matter how strange the distribution of the walls or cameras is. As detailed in Section 3.6, the calibration of the environment for cloud fusion is only needed once, and then it is valid for every execution of the system. As detailed in Section 3.7, the algorithm for skeleton merging is robust even though several Kinects are generating different skeletons for the same person in the scene.

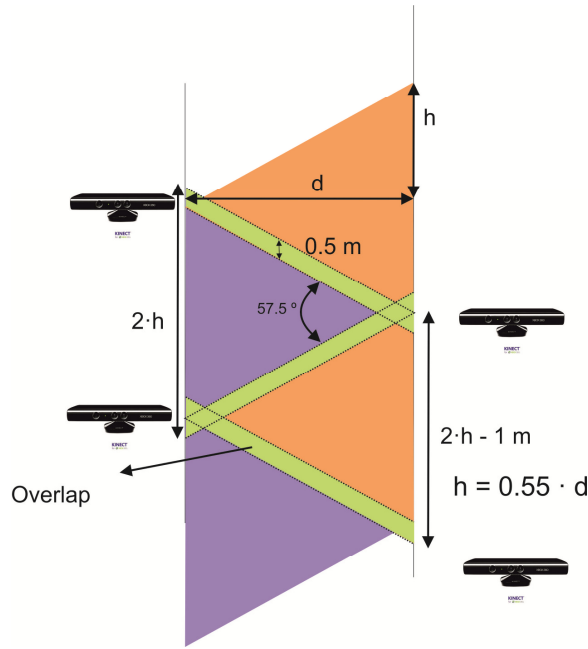

Figure 4. Final Kinect cameras layout with an overlap of 0.5 m

### 3.3 Data acquisition

To add more Kinect devices to the scene, the naïve solution is to try to connect multiple cameras to the same computer. However, due to the large volume of data generated by each camera, a USB controller is needed to handle the bandwidth emitted by each one. As a consequence, that is not a valid solution, since most computers only support a limited number of USB controllers, usually two or three. The solution adopted for our system was to develop a distributed application with multiple cameras connected to multiple computers.

The data provided by each Kinect in which we are interested in are: a three-channel *RGB image* of the scene, captured at a resolution of 640x480 px; a *depth map*, which is a texture of the same resolution in which each pixel takes a value that indicates the distance between the infrared pattern and the sensor; a texture of *user labels* with same resolution, in which each pixel takes the value of the user id in front of the camera or a zero value; and the *skeletons* of the users in the scene, which are formed by joints representing the parts of the body (head, elbow, shoulders …) and include both the xyz position and rotation from an initial pose position.

### 3.4 Data coding

Before the information captured by the remote devices can be sent, it has to be encoded. Kinect data to be processed includes RGB images, depth maps, user labels, and skeleton joints. Skeleton joints information is sent without any compression, since the volume of data needed to store and transmit the position of all the joints is negligible compared to the volume of image or depth information.

To encode the RGB image we use a video compressor. It would be meaningless to use an image codec such as JPEG, since it only uses spatial information at the time of compression and thus the needed bit rate is much higher. For video compression, the proposed system uses the cross-platform library FFMPEG, which provides many audio and video codecs. The RGB image compression is done using the VP8 codec developed by Google TM [19] that needs a YUV 420

The International Journal of Multimedia & Its Applications (IJMA) Vol.6, No.1, February 2014

image format. Although VP8 codec introduces quality losses during compression and decompression, its balance between final quality and performance makes it adequate for our purposes. Additionally, the loss in quality remains quite low and the human eyes, acting as filters, are not able to appreciate it.

As it has been commented before, there is no compression codec to encode depth or a combination of RGB and depth information. In the proposed system, the compression of the depth map has to be done in a tricky way, based on the scheme proposed by Pece et al. [20]. Basically, one depth channel has to be converted into a three-channel image, and then a specific codec H264 is used to compress the result. The H264 codec is more computationally expensive than VP8 and it also needs more bandwidth. However, the final results obtained for the particular case of depth-information are much better than using VP8.

Finally, for labels codification, the chosen codec was VP8. Using this codec, quality losses which could result in user misidentifications in the remote computer, can be expected. In order to prevent these situations, the following strategy is proposed. Since encoders usually join together colors being too close, we propose spacing them before codification. The values 0 to 15 of user labels are translated into values from 0 to 255, preventing the encoder from mixing them up. Figure 5 shows the employed conversion equivalences. With these new values, labels are stored into a luminance channel and then compressed.

| 0 | 1 | 2 | 3 | 4 | 5 | 6 | 7 | 8 | 9 | 10 | 11 | 12 | 13 | 14 | 15 |
|---|---|---|---|---|---|---|---|---|---|----|----|----|----|----|----|
| 0 | 17 | 34 | 51 | 68 | 85 | 102 | 119 | 136 | 153 | 170 | 187 | 204 | 221 | 238 | 255 |

Figure 5. Correspondence of user labels to colors to avoid misidentifications after data compression.

Due to the computation requirements of the proposed system, in order to code and send the 3D information, it has been designed to process video sequences in parallel using threads. The RGB image, depth map, user labels and skeletons are acquired at the same time. Each type of data is then coded separately in parallel using the 'Boost' threads library.

### 3.5 Data transmission

System servers are registered in a CORBA naming service after starting, so that the system's central computer can find them, without needing to know their IP addresses. When the central computer establishes a communication and asks for data, the server collects the information from every local attached Kinect, encodes it and sends it continuously to the remote client. The client is constantly receiving data sent from each server, but it may not use all the information that arrives to the client. The system is designed to decode only the information that is to be used. To this end, mutual exclusion techniques are employed.

Compressed information is stored into CORBA data arrays. Then, the server sends data by invoking remote methods in each client. These methods receive input arguments containing the compressed RGB image, depth map, user labels, and uncompressed skeletons. The information is sent only to the clients who have previously registered on the server.

Each camera sends a volume of data $D$ of about 70.000 bytes. The theoretical bandwidth needed to receive this data at a frame rate $F$ from every camera, depending on the number of cameras $N$, can be computed according to equation (2).

$$BW(bps) = 8NDF \qquad (2)$$

The International Journal of Multimedia & Its Applications (IJMA) Vol.6, No.1, February 2014

Table 1 shows the bandwidth needed by the system for data transmission at a frame rate F=30 fps, depending on the number of cameras.

| No. Cameras | Bandwidth |
|---|---|
| 1 | 16.8 Mbps |
| 2 | 33.6 Mbps |
| 3 | 50.4 Mbps |
| 4 | 67.2 Mbps |
| 5 | 84 Mbps |

Table 1. Bandwidth consumption for 30fps data transmission, depending on number of Kinects.

## 3.6 Point cloud fusion

Once the Kinect cameras have been installed in the location to be monitored, a first system calibration has to be performed. The goal of the calibration is that the central client computer can find the proper transformation matrices to align and fuse the received point clouds. One of devices is chosen to be the center of the coordinate system and then rotations are calculated from the other cameras. Given each pair of point clouds, the objective is to calculate the 4x4 rotation and translation matrices by solving the system of equations **B = RA + t**, where **A** and **B** are three-component points, **R** is a 3x3 rotation matrix and **t** is a three-component column translation vector.

Within the system interface, the calibration step will prompt the user to check the correspondence of at least 3 common points in different clouds of points. This calibration clouds are not yet compressed for better results. For this purpose, it is useful to place an object into the intersection area of different infrared patterns. Figure 6 shows this process using points belonging to a chair and a box on top of it. Taking the marked common points, the system can approximate an initial calibration, which serves to rotate the point clouds and apply the algorithm ICP (Iterative Closest Point), which refines the calibration. These rotation matrices have only to be computed the first time the system is deployed and they are later used to rotate all information coming from the different cameras, including user skeletons, in different executions.

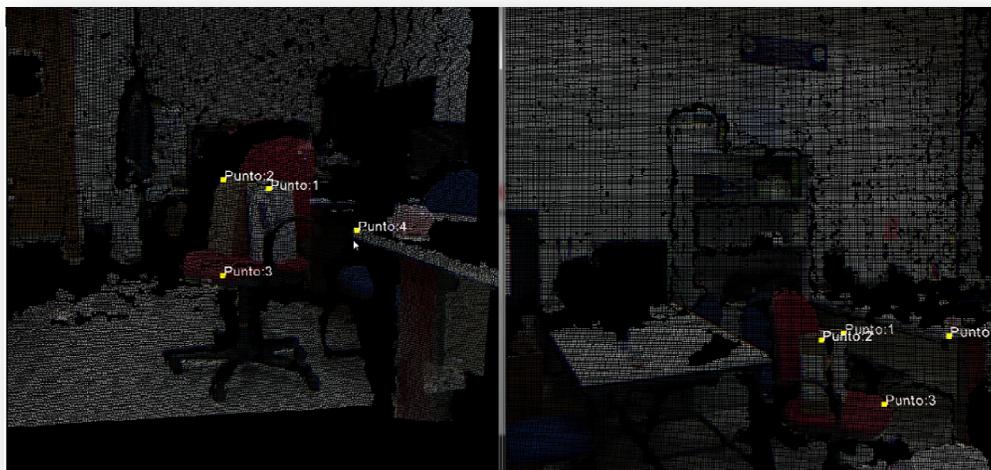

Figure 6. Initial calibration to determine rotation and translation matrices. These matrices are used to fuse 3D information coming from different Kinect cameras.

69

The International Journal of Multimedia & Its Applications (IJMA) Vol.6, No.1, February 2014

## 3.7 Skeleton merging

We distinguish between Kinect input skeletons and system's final output skeletons. Each output skeleton is computed dynamically from a linked list of input skeletons, which are merged and averaged together in the client machine.

Figure 7 depicts the process of output skeletons computation. To merge the input skeletons, the first step is to apply rotation matrices to the detected joints. Once all the skeletons from all the cameras are in the same reference system, the algorithm for skeleton merging can be applied. The first step is to check for changes in the previous linked lists of skeletons, which contain the correspondence among similar input skeletons. These lists include the camera identifiers, the input skeletons identifiers and the output skeletons identifiers. Every time a camera provides information of a new skeleton, the system tries to add it into a linked list. The estimated position of the person in a camera is complemented taking into account the information of speed (meters per frame) and walk direction. This information is updated every frame by collecting the information of its center of mass in a circular buffer, where old values are overwritten in every frame. Additionally, evaluating skeletons matching during 15 consecutive frames strengthens the robustness of the system. The conditions that have to be satisfied so that two input skeletons, detected by two different Kinects, can be merged are:

1. The centers of masses of both input skeletons should be at a distance smaller than a given threshold (i.e. 15 cm). A similar strategy has been used in other human skeleton tracking proposals [21].
2. The speed vectors associated to both input skeletons should be similar enough (i.e. the angle between vectors close to 0) and should have the same magnitude.

The algorithm is robust for skeleton merging, even though a skeleton is detected from several Kinect cameras. The skeleton tracking keeps working although the person is standing in a position that is covered by more than two or three cameras. In case no correspondence can be found to include the new skeleton into any existing linked list, then it is considered as a new output skeleton and a new linked list is built up. The final joints are calculated by averaging all the joints from different cameras. If the confidence of a specific joint within an input skeleton is smaller than 0.5, that particular joint will not contribute to calculate the final joint in the output skeleton. The advantage of this design is that in case one camera cannot detect a given joint, its position can be determined from the information given by other camera. The probability of having all the joints describing the available skeleton, and with accurate positions, is increased with the number of cameras detecting the skeleton.

If the room allows a layout of Kinect devices similar to the one depicted in Figure 4, then the skeleton tracking algorithm takes advantage of some previous knowledge of the cameras distribution. Basically, the system knows precisely the order in which the cameras are placed, thus can anticipate the event of a person leaving one camera and walking into the next camera's field of view.

The first step to improve the algorithm is to check the skeletons already covered by a given camera. The information of those skeletons that are yet in the scene is updated in every frame, while the last information of those skeletons that are no longer detected is kept in memory for the next 15 frames. This is the time needed to determine if that skeleton has abandoned the place or if the person has just changed his or her position from the last camera to the previous or next camera in the layout. The position where a new skeleton is detected in either of these cameras is compared against the estimated position of the previously lost skeleton, so that we can match skeletons and labeled information about the user is not lost. The new condition that has to be satisfied so that a lost skeleton and a new detected skeleton can be merged is:

1. There is a correspondence between the boundaries of the cameras' field of view. When a person disappears from one camera by the left side, he or she must appear into the field of view of an opposite camera also by the left side.

If the three conditions stated above are met, the algorithm merges the involved skeletons. They represent indeed the same person, so critical information (i.e. label) about the lost skeleton has to be attached to the new skeleton.

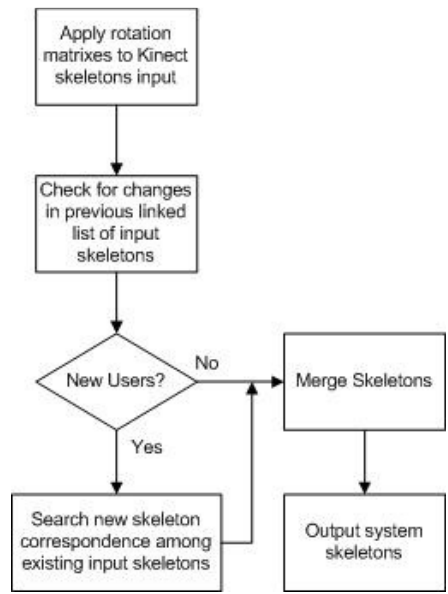

Figure 7. Skeleton merging algorithm.

## 3.7 Data visualization and skeleton tracking

The central client interface uses PCL to visualize the final reconstructed space and allows real-time tracking of labeled people inside the area covered by the cameras. Figure 8 shows a labeled skeleton being monitorized in real-time by 5 cameras. The final scene can be rotated and analyzed from any point of view. However, there is a limitation on the available frame rate due to the VTK rendering methods employed by the system. When the number of points in the final point cloud grows, the frame rate is reduced. This is not a problem related to the compression/decompression computational cost, but the visualization methods included in PCL. Future releases of PCL are said to address this problem by adding native OpenGL rendering [17]. In order to guarantee usability of the system, despite of this problem, the user interface allows subsampling the number of points to visualize. Test and results section gives some figures of performance with 5 device cameras.

Finally, the system is designed to store skeleton information of people within the tracking area, associated to their labeled output skeletons. Once recording has started, all the user joints are stored in a raw file that can be further used to reproduce any occurred situation, or serve as an input for another application for situation analysis (i.e. movement recognition application). For applications that require human activity registration, the needed storage space is much smaller than in conventional 2D video systems, since only the skeletons may need to be stored.

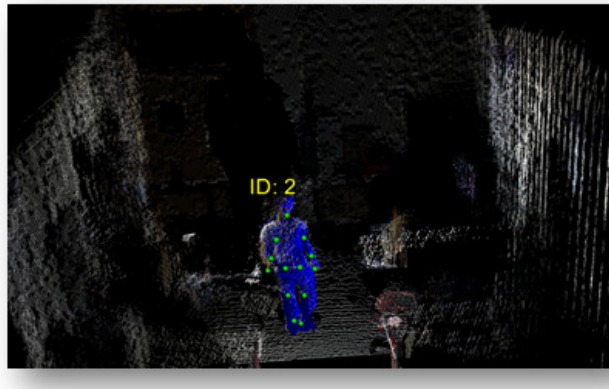

Figure 8. Labeled skeleton and associated joints obtained from the combination of 5 cameras information.

## 4. TESTS AND RESULTS

The system has been tested using 5 Kinect cameras connected to three personal computers: two desktop computers equipped with an Intel i5-2400 CPU running at 3.10 GHz and a laptop equipped with an Intel i7-3610QM CPU running at 2.30 GHz. The client computer was an Intel Xeon X5650 with 24 cores running at 2.66 GHz and equipped with an Nvidia GeForce GTX580 GPU. The server computers were running Windows 7 operating system, while the central client was running Linux Fedora 16.

The aim of these tests was to check the performance of the final system in real conditions. The first conducted tests included data transmission, coding/decoding and visualization measurements. Using RGB input at 640x480 px resolution and coding depth information to 320x240 px color maps, the theoretical limit on the number of cameras that can be connected over a Gigabit Ethernet is higher than 50 for a frame rate of 30 fps. These numbers do not consider the overhead of TCP connection. In our experimental tests, performed with up to 5 cameras (the maximum number of cameras we managed to have), the obtained frame rate was actually 30 fps. However, in our tests, we detected that the final scene rendering was affected by VTK visualization limitations of PCL, even though every server was transmitting at 30 fps and the client computer was decoding all cameras information at the same frame rate. As explained above, the achieved frame rate depends on the number of points in the cloud. Table 2 shows how visualization of a cloud of points constructed from 5 cameras renders only at 7 fps if every point is drawn onto the screen. Subsampling the number of points by 16, which actually still provides a very nice representation of the scene, improves performance to 29 fps.

| Frame rate (fps) | Rendered points |
|---|---|
| 7 | $\approx 5*307200 = 1.536.000$ |
| 13 | $\approx 5*307200/4 = 384000$ |
| 22 | $\approx 5*307200/9 = 170666$ |
| 29 | $\approx 5*307200/16 = 96000$ |

Table 2. Frame rate obtained during visualization using VTK for 5 cameras 3D reconstruction. This is a limitation of VTK, not the system itself.

The second battery of tests conducted included situations to measure the behavior of the system with different people in the scene and measure the robustness to self-user and inter-user occlusions. The first test consisted in a user placed in the center of the scene. Meanwhile, another user revolves around him or her, so that some cameras can see the first user and some others cannot. The goal is to test the robustness of the system when different cameras detect and lose Kinect skeletons over and over again. The test was conducted ten times using combinations of different users' height and the obtained result was always successful in every situation, since the system did not confuse users or incorrectly merged their skeletons. The second test consisted in users sitting and getting up from chairs in an office space. This test measured the robustness of the system to some skeletons joint occlusions, since some of the cameras are not able to provide accurate positions for body parts behind tables or chairs. The test was repeated for ten different people sitting in front of the four tables in the scene in Figure 9 and again the system worked perfectly. The third test consisted in covering and uncovering one by one the different cameras in the scene while 5 people were being tracked in the scene. The goal was to test what happens when multiple Kinect input skeletons are removed and detected at the same time. The result was again satisfactory and every computed output skeletons in the scene kept being tracked consistently.

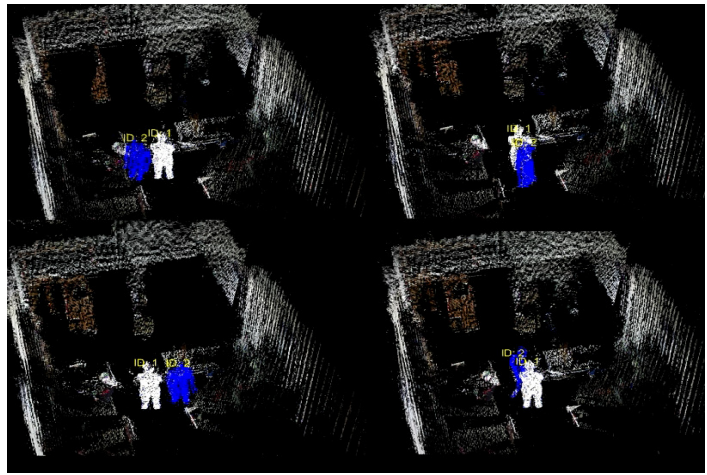

Figure 9. Inter-user occlusion test within a space monitorized by 5 cameras.

## 5. CONCLUSIONS

This article describes a distributed CORBA system for remote space 3D monitoring using Kinect consumer depth-cameras. Due to the high bandwidth needs of these cameras, the maximum number of cameras that can be connected to a single computer is usually two. The solution provided in this paper includes a client-server application that can handle the information acquired by any number of cameras connected to any number of computer servers. Since one Kinect camera can only detect precisely the depth information within a field of view of 3.5 meters, the proposed system solves, at the same time, the limitation on the size of the location that can be monitored precisely. A central client computer can be used to monitor the reconstructed 3D space in real time and track the movements of people within it.

In the central client computer, a skeleton-merging algorithm is used to combine the information of skeletons belonging to the same person, but generated by different Kinect cameras, into a single output skeleton. The conducted tests showed that this algorithm is robust under several situations, avoiding unwanted duplication of skeletons when new people enter the scene or under camera or inter-user occlusions. Moreover, the algorithm combines the information coming from each skeleton joint independently, so the 3D location of joints in the final generated skeleton is more precise, having been averaged among all the cameras that detected that joint. In case a self-user or a inter-user occlusion causes one joint not to be detected by one or more of the cameras, its position is reconstructed using the information coming from cameras in which the joint has been detected with enough confidence. Output skeleton movements can be stored in raw files for further analysis of situations.

This system provides a very precise and convenient way of monitoring a 3D space at an affordable price. People activity in the scene can be registered for further analysis and the storage needed to keep track of human behavior under different circumstances can be much lower than for conventional 2D systems, if only the skeletons are needed. Future research tasks will include designing a top activity recognition layer that could monitor people behavior and interactions.

## ACKNOWLEDGEMENTS

This research was supported in part by the Ministerio de Ciencia e Innovación under project TIN2010-20529 and Junta de Castilla y León under project VA171A11-2.

## REFERENCES

[1] Sage, K. & Young S. (1999) Security Applications of Computer Vision, Aerospace and Electronic Systems Magazine, IEEE 14(4):19-29.
[2] Boutaina, H., Rachid, O.H.T. & Mohammed, E.R.T. (2013) Tracking multiple people in real time based on their trajectory, in Proc. of Intelligent Systems: Theories and Applications (SITA), 8th International Conference on, pp.1-5, Rabat, Morocco.
[3] Strbac, M., Markoviu, M., Rigolin, L. & Popoviu, D.B. (2012) Kinect in Neurorehabilitation: Computer vision system for real time and object detection and instance estimation, in Proc. Neural Network Applications in Electrical Engineering (NEUREL), 11th Symposium on, pp. 127-132, Belgrade, Serbia.
[4] Martín Moreno, J., Ruiz Fernandez, D., Soriano Paya, A. & Berenguer Miralles, V. (2008) Monitoring 3D movements for the rehabilitation of joints in Physiotherapy, in Proc. Engineering in Medicine and Biology Society (EMBS), 30th Annual International Conference of the IEEE, pp. 4836-4839, Vancouver, Canada.
[5] Ukita, N. & Matsuyama, T. (2002) Real-Time Cooperative Multi-Target Tracking by Communicating Active Vision Agents, in Proc. Pattern Recognition, 16th International Conference on, vol.2, pp.14-19.
[6] R Bodor, R., Jackson, B. & Papanikolopoulos, N. (2003) Vision-Based Human Tracking and Activity Recognition, in Proc. of the 11th Mediterranean Conf. on Control and Automation, pp. 18-20.
[7] Poppe R. (2007). Vision-based human motion analysis: An overview, Computer Vision and Image Understanding 108:4–18.
[8] Weinland, D., Ronfard, R. & Boyer, E. (2011) A survey of vision-based methods for action representation, segmentation and recognition, Computer Vision and Image Understanding 115(2):224–241.
[9] Kinect for Windows (2013) Kinect for Windows. Retrieved from http://www.microsoft.com/en-us/kinectforwindows/develop/overview.aspx, last visited on July 2013.
[10] Burrus, N. (2013) RGBDemo. Retrieved from http://labs.manctl.com/rgbdemo/, last visited on July 2013.
[11] KinectTCP (2013) KinectTCP. Retrieved from https://sites.google.com/a/temple.edu/kinecttcp/, last visited on July 2013
[12] OpenNI (2013) OpenNI. Plataform to promote interoperability among devices, applications and Natural Interaction (NI) middleware. Retrieved from http://www.openni.org, last visited on July 2013.
[13] Prime Sense (2013) Prime Sense. Retrieved from http://www.primesense.com/, last visited June 2013.
[14] Martínez Rach, M.O., Piñol, P., López Granado, O. & Malumbres, M.P. (2012) Fast zerotree wavelet depth map encoder for very low bitrate, in Actas de las XXIII Jornadas de Paralelismo, Elche, Spain.
[15] Joon-Heup, K., Moon-Sang J. & Jong-Tae, P. (2001) An Efficient Naming Service for CORBA-based Network Management, in Integrated Network Management Proceedings, IEEE/IFIP International Symposium on, pp.765-778, Seattle, USA

[16] PCL (2013) Point Cloud Library. Retrieved from http://pointclouds.org/, last visited on July 2013.
[17] PCL Developers Blog (2013) PCL Developers Blog. Retrieved from http://pointclouds.org/blog/, last visited on July 2013.
[18] Parajuli, M., Tran, D.; Wanli, Ma; Sharma, D. (2012) Senior health monitoring using Kinect, in Proc. Communications and Electronics (ICCE), Fourth International Conference on, pp. 309-312, Hue, Vietnam.
[19] The WebM Project (2013) The WebM Project. Retrieved from http://www.webmproject.org, last visited on July 2013.
[20] Pece, F., Kautz, J. & Weyrich, T. (2011) Adapting Standard Video Codecs for Depth Streaming, in Proc. of the 17th Eurographics conference on Virtual Environments & Third Joint Virtual Reality (EGVE - JVRC), Sabine Coquillart, Anthony Steed, and Greg Welch (Eds.), pp. 59-66, Aire-la-Ville, Switzerland.
[21] García, J., Gardel, A., Bravo I., Lázaro, J.L., Martínez, M. & Rodríguez, D. (2013). Directional People Counter Based on Head Tracking, IEEE Transactions on Industrial Electronics 60(9): 3991-4000.

## AUTHORS

Mario Martínez-Zarzuela was born in Valladolid, Spain; in1979. He received the M.S. and Ph.D. degrees in telecommunication engineering from the University of Valladolid, Spain, in 2004 and 2009, respectively. Since 2005 he has been an assistant professor in the School of Telecommunication Engineering and a researcher in the Imaging & Telematics Group of the Department of Signal Theory, Communications and Telematics Engineering. His research interests include parallel processing on GPUs, computer vision, artificial intelligence, augmented and virtual reality, and natural human-computer interfaces.

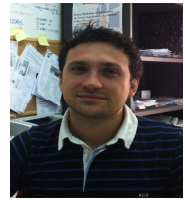

Miguel Pedraza Hueso was born in Salamanca, Spain, in 1990, He received his title in Telecommunication Engineering from the University of Valladolid, Spain, in 2013. Since 2011, he has collaborated with Imaging & Telematics Group of the Department of Signal Theory, Communications and Telematics Engineering. His research interests include applications with Kinect and augmented reality.

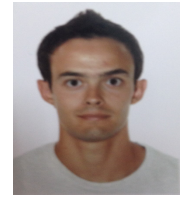

Francisco Javier Díaz-Pernas was born in Burgos, Spain, in1962. He received the Ph.D. degree in industrial engineering from Valladolid University, Valladolid, Spain, in 1993. From 1988 to 1995, he joined the Department of System Engineering and Automatics, Valladolid University, Spain, where he has worked in artificial vision systems for industry applications as quality control for manufacturing. Since1996, he has been a professor in the School of Telecommunication Engineering and a Senior Researcher in Imaging & Telematics Group of the Department of Signal Theory, Communications, and Telematics Engineering. His main research interests are applications on the Web, intelligent transportation system, and neural networks for artificial vision.

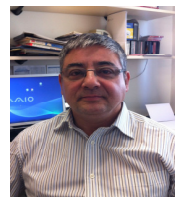

David González-Ortega was born in Burgos, Spain, in 1972. He received his M.S. and Ph.D. degrees in telecommunication engineering from the University of Valladolid, Spain, in 2002 and 2009, respectively. Since 2003 he has been a researcher in the Imaging and Telematics Group of the Department of Signal Theory, Communications and Telematics Engineering. Since 2005, he has been an assistant professor in the School of Telecommunication Engineering, University of Valladolid. His research interests include computer vision, image analysis, pattern recognition, neural networks, and real-time applications.

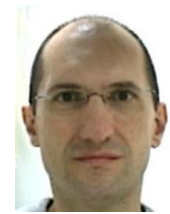

Míriam Antón-Rodríguez was born in Zamora, Spain, in 1976. She received her M.S. and Ph.D. degrees in telecommunication engineering from the University of Valladolid, Spain, in 2003 and 2008, respectively. Since 2004, she is an assistant professor in the School of Telecommunication Engineering and a researcher in the Imaging & Telematics Group of the Department of Signal Theory, Communications and Telematics Engineering. Her teaching and research interests includes applications on the Web and mobile apps, bio-inspired algorithms for data mining, and neural networks for artificial vision.

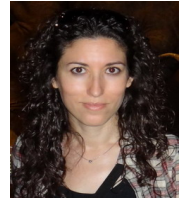